\documentclass[final]{cvpr}

\usepackage{times}
\usepackage{epsfig}
\usepackage{graphicx}
\usepackage{amsmath}
\usepackage{amssymb}
\usepackage{multirow}

\usepackage[pagebackref=true,breaklinks=true,colorlinks,bookmarks=false]{hyperref}

\def\cvprPaperID{****} 
\def\confYear{CVPR 2021}

\begin{document}

\title{Video Based Fall Detection Using Human Poses}

\author{Ziwei Chen\\
Southeast University\\
Nanjing, China\\
{\tt\small richard\_chen@seu.edu.cn}
\and
Yiye Wang\\
Southeast University\\
Nanjing, China\\
{\tt\small 230208761@seu.edu.cn}

\and
Wankou Yang\\
Southeast University\\
Nanjing, China\\
{\tt\small wkyang@seu.edu.cn}
}

\maketitle

\begin{abstract}
Video based fall detection accuracy has been largely improved due to the recent progress on deep convolutional neural networks. However, there still exists some challenges, such as lighting variation, complex background, which degrade the accuracy and generalization ability of these approaches. Meanwhile, large computation cost limits the application of existing fall detection approaches. To alleviate these problems, a video based fall detection approach using human poses is proposed in this paper. First, a lightweight pose estimator extracts 2D poses from video sequences and then 2D poses are lifted to 3D poses. Second, we introduce a robust fall detection network to recognize fall events using estimated 3D poses, which increases respective filed and maintains low computation cost by dilated convolutions. The experimental results show that the proposed fall detection approach achieves a high accuracy of 99.83\% on large benchmark action recognition dataset NTU RGB+D and real-time performance of 18 FPS on a non-GPU platform and 63 FPS on a GPU platform. 
\end{abstract}

\begin{figure*}[ht]
\begin{center}
\includegraphics[scale=0.6]{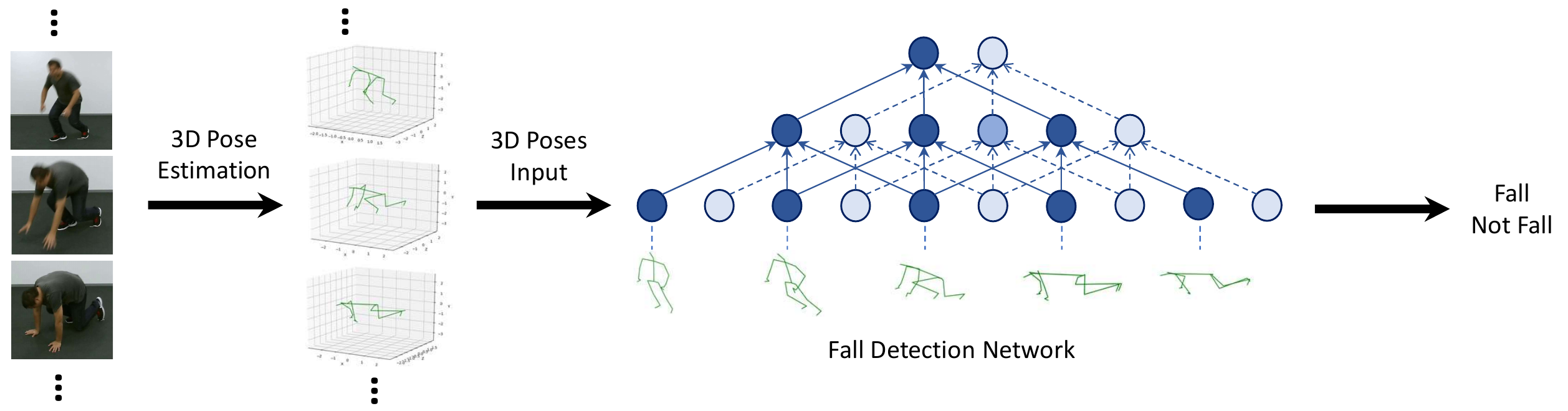}
\end{center}
\caption{Overall structure of our fall detection approach. Video sequences are first sent into pose estimator to get 3D poses, then fall detection network takes 3D poses to classify action class.}
\label{Pipeline}
\end{figure*}

\section{Introduction}

Nowadays, the ageing of the population has become a global phenomena, there were 727 million persons aged 65 or over in 2020 and the number of the elderly worldwide will be projected to more than double over the next three decades, sharing around 16.0 per cent of the population in 2050 \cite{UN}. According to the World Health Organization (WHO) \cite{WHO}, adults older than 65 years suffer the greatest number of fatal falls, which could cause serious injuries and even death. Therefore, intelligent fall detection has drawn increasing attention from both academia and industry and has become an urgent need for vulnerable people, especially the elderly.

The existing fall detection methods can be roughly divided into two categories, which are wearable sensor based methods and vision based methods \cite{review1}. Wearable sensors, including accelerometer, gyroscopes, pressure sensor and microphone, can detect the location change or acceleration change of human body for fall detection. However, inconvenience is still the main problem that many elderly people are unwilling to wear sensors all day. Besides, wearable sensors may be affected by noise and some daily activities like lying or sitting on the sofa quickly may lead to false alarm. With the rapid development of computer vision and deep learning techniques in recent years, the number of proposed vision based methods has increased a lot \cite{I3}. Compared with wearable sensor based methods, vision based methods are free from the inconvenience of wearing the device. While the detection accuracy of vision based methods has increased a lot in recent years, false detection still may occur as a result of lighting variation, complex background and so on. Furthermore, vision based methods, especially deep learning based methods, have a large computation cost which makes it hard to achieve real-time performance. To conclude, how to maintain a high detection accuracy while lower the computation cost is a valuable research topic.

Human pose estimation is a fundamental task in computer vision, the goal is to localize human keypoints in a image or 3D space. Human pose estimation has many applications, including human action recognition, human-computer interaction, animation, etc. Nowadays, 2D human pose estimation \cite{I4,I5,I6,I7,transpose} has achieved convincing performance on some large public datasets \cite{I8,I9} and the performance of 3D human pose \cite{I10,I11,I12,I13} estimator has been improved a lot. Using human poses can alleviate the problem of lighting variation or complex background in fall detection task, so as to effectively improve the accuracy and generalization ability of fall detection methods.

Although existing vision based methods have brought fall detection accuracy to such a high level, they rely on a large computation \cite{I14,I15,I16} and features extracted for classification are not robust for some challenging conditions. Meanwhile, the period of falling differs among people \cite{I17}, the fall of the elderly lasts longer than other groups and some daily activities such as lying to bed differ from fall. So more video frames should be taken as the input. However, previous works \cite{I18,I19,F8} only take a short video sequence as the input which may lead to the false detection.

To address the above problems, we propose a video based fall detection approach using human poses in this paper. Our fall detection approach consists of two steps: (1) estimating 3D poses in video sequences (2) recognize fall events from estimated 3D poses. The first step is compatible with any state-of-the-art pose estimator. We formulate 3D pose estimation as 2D pose estimation followed by 2D-to-3D pose lifting. A lightweight 2D pose estimator and a lifting network are adopt to lower the computation cost. At the second step, we present a fall detection network taking 3D poses of each frame as input. In order to achieve a convincing accuracy while maintaining the low computation cost for long video sequences, one-dimensional dilated temporal convolution \cite{I20} is adopt. 

In summary, this paper has three contributions:
\begin{itemize}
\item[$\bullet$] We propose a fall detection model, which includes a 3D pose estimator and a fall detection network based on human poses.
\item[$\bullet$] We explore the effects of factors which could contribute to the performance of fall detection including input joints, loss function.
\item[$\bullet$] Our approach achieves a high accuracy at 99.83\% on NTU RGB+D dataset and real-time performance on non-GPU platform.
\end{itemize}

\section{Related Work}

The related work on vision based fall detection are first reviewed and the difference between them and our work are discussed. Then we review some work on human pose estimation.

    \subsection{Vision Based Fall Detection}

    Many approaches have been proposed for vision based fall detection \cite{F1,F2,F3,F4,F5,F6,F7,F8,F9}. These approaches differ in terms of the used sensors and classify algorithms.

    Sensors for most vision based approaches are RGB cameras, depth cameras, infrared cameras and Kinect. In \cite{F1}, Lu et al. propose to detect fall on RGB videos using 3D CNN combined with LSTM to address the problem of insufficient fall data. Shojarei et al \cite{F7}. obtain 3D coordinates of human keypoints using depth camera and then do fall detection based on these poses. Zhong et al. \cite{F9} propose an end to end solution within a multi-occupancy living environment by thermal sensors. Nowadays, the use of Kinect for fall detection \cite{F5,F6} has been increased a lot as 3D information can be obtained. However, depth camera in Kinect has a restricted distance which makes it unsuitable for a large space. 

    Decision trees, SVM and threshold are used to classify fall action categories \cite{F10,F11,F12,F13}. Compared with these algorithms, deep neural network can achieve higher classification accuracy and avoiding feature engineering task. Adhikari et al. \cite{F14} propose a fall detection system using CNN to recognize Activities of Daily Life (ADL) and fall events. A 3D CNN is developed in \cite{F15} to improve fall detection accuracy by exploring spatial and temporal information. In \cite{F16}, Variational Auto-encoder (VAE) with 3D convolutional residual block and a region extraction technique for enhancing accuracy are used to detect fall actions.

    Our approach is similar to Tsai et al. \cite{F8}, thus we provide more detailed comparison. Tsai et al. \cite{F8} propose a traditional algorithm to transform depth information into 3D poses and use 1D CNN to detect fall events. In \cite{F8}, depth camera is used while our human pose estimator can obtain 3D poses directly from RGB images which makes it free from limited measurement distance. Though 1D CNN is used in both works, only 30 frames are taken as input in their approach while our model can take 300 frames at most as input. Some actions last longer and the elderly fall slower than the young so it is necessary to recognize fall events using long video sequences.

    \subsection{Human Pose Estimation}

    Human Pose Estimation is to localize human keypoints, which can be categorized as 2D and 3D human pose estimation according to the output.

    Deep learning has become a promising technique in 2D human pose estimation in recent years. Firstly, CNN is introduced to solve 2D pose estimation problem by directly regressing the joint coordinates in DeepPose \cite{HPE1}. Then joint heatmaps \cite{HPE2} have been widely adopt as training signals in 2D human pose estimation for great performance. Newell et al. \cite{HPE3} propose an U-shape network by stacking up several hourglass modules to refine prediction. The work by Cao et al. \cite{I7} detects all human keypoints first and assembles them to different person by part affinity fields (PAFs), which achieves real-time performance in multi person pose estimation task. HRNet \cite{I6} adapts the top-down pipeline and generates convincing performance via maintaining the high resolution of feature maps and multi-scale fusion. 

    3D human pose estimation is to localize the position of human keypoints in 3D space from images or videos. The early 3D human pose estimation methods directly predict the 3D joint coordinates via deep neural networks \cite{HPE4}. While a set of features suitable for the task can be spontaneously learned, these models usually have large computation cost and high complexity. Due to the development of 2D human pose estimation, methods based on 2D poses \cite{HPE5,HPE6,HPE7} have become the main stream. 2D poses are concatenated as the input to predict the depth information of each keypoint, greatly reducing the complexity of the model. In order to overcome the problem of insufficient data in 3D human pose estimation, wandt et al. \cite{HPE7} apply weak supervised learning to 3D human pose estimation by using reprojection method to train 3D mapping model. Cheng et al. \cite{HPE8} explore video information to further modify the human poses to avoid incorrect pose estimation. He et al. \cite{HPE9} use multi view images to overcome the occlusion problem based on epipolar geometry and achieve great performance.

\section{Video Based Fall Detection}
The goal of this work is to establish a video based fall detection approach using human poses. The general framework of the proposed approach is shown in Figure \ref{Pipeline}. Firstly, human pose estimator is applied to each video frame to generate a set of human poses. Then fall detection network works to recognize whether there is a fall event based on human poses.

\begin{figure}[ht]
\begin{center}
\includegraphics[scale=0.6]{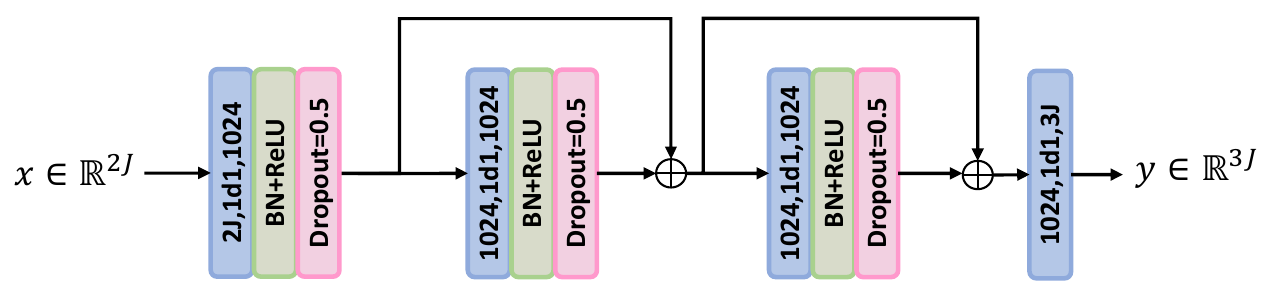}
\end{center}
\caption{Lifting network. "$2J$" means 2D coordinates of $J$ joints are concatenated.}
\label{lifting}
\end{figure}

\subsection{Human Pose Estimator}
Our fall detection approach do not rely on any specific pose estimator. When obtaining 3D poses, we follow the widely-used pipeline in 3D human pose estimation \cite{HPE6}, which predicts 2D human poses in the first step and lifts 2D poses to 3D poses. For 2D human pose estimation, we adapt off-the-shelf Lightweight Pose Network (LPN) \cite{LPN} considering its low complexity and adequate accuracy. LPN is pretrained on MS COCO \cite{coco} dataset and fine tuned on NTU RGB+D dataset \cite{ntu} for our task.

Lifting network takes 2D human poses $x\in \mathbb{R}^{2J}$ as input and lifts them to 3D poses $y\in \mathbb{R}^{3J}$. The goal is to find a function $f^*: \mathbb{R}^{2J} \xrightarrow[]{} \mathbb{R}^{3J}$ that minimizes the prediction error of $N$ poses over a dataset: \\
\begin{equation}
    f^* = \min_{f} \frac{1}{N} \sum_{i=1}^{N} \mathcal{L}(f(x_{i})-y_{i})
\end{equation}

Figure \ref{lifting} shows the structure of our lifting network, it consists of two residual blocks. 2D coordinates of joints are concatenated so one-dimensional convolution can be adopt to reduce parameters and complexity. The core idea of our lifting network is to predict depth information of keypoints effectively and efficiently. Note that what we care is not the absolute location of keypoints in 3D space but the relative location between them. So before a 2D pose is lifted to 3D, it is normalized by centering to its root joint and scaled by dividing it by its Frobenius norm. We define the prediction error as the squared difference between the prediction and the ground-truth pose:
\begin{equation}
    \mathcal{L}(\hat{y_{i}}, y_{i}) = ||\hat{y_{i}} - y_{i}||_{2}^{2}
\end{equation}
where $y_{i}$ and $y_{i}$ are estimated and the ground-truth relative position of the $i$-th pose.

\begin{figure}[ht]
\begin{center}
\includegraphics[scale=1.0]{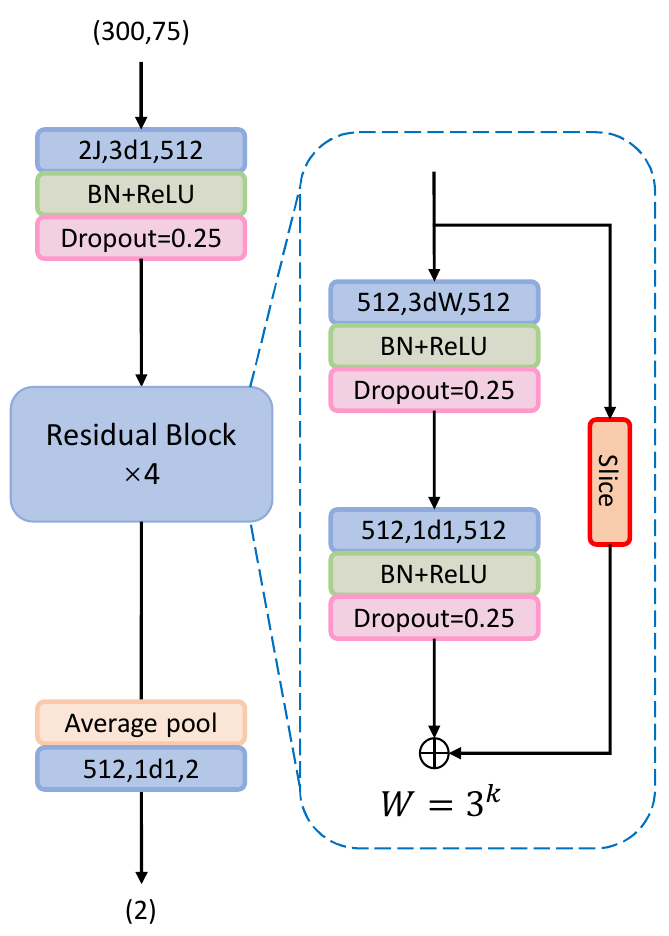}
\end{center}
\caption{Fall detection network.}
\label{TCN}
\end{figure}

\subsection{Fall Detection Network}
Our fall detection network is a fully convolutional architecture with residual connections that takes a sequence of 3D poses $X\in \mathbb{R}^{T\times 3J}$ as input where $T$ is the number of frames and predict whether there is a fall behavior. In convolutional networks, the path of gradient between output and input has a fixed length, which mitigates vanishing and exploding gradients. It is important for our task as $T$ was set to 300 to recongize falls in such a long video sequence. Moreover, dilated convolutions are applied in our network to model long-term dependencies while maintaining efficiency.

When 3D poses are obtained by pose estimator, we also do the centering and scaling. Besides, 3D poses are firstly rotated by paralleling the bone between hip and spine to the $z$ axis, then by paralleling the bone between left shoulder and right shoulder to the $x$ axis, normalized 3D poses can be obtained. 

Figure \ref{TCN} shows our fall detection network. 3D coordinates $(x,y,z)$ of $J$ joints for each frame are concatenated as the network input and a convolution with kernel size 3 and $C$ output channels is applied. 
This is followed by $N$ residual blocks. Each block includes two one-dimensional convolutions, the first one is dilated and dilation factor is $W=3^N$, followed by another 1D convolution with kernel size 1. Batch normalization, rectified linear units and dropout are used after every convolution except the last one. With dilation factor, each block increases the receptive field to exploit temporal information without too much computation increasement. We use unpadded convolutions so the output size of each block is different, details can be seen in Table \ref{size}. Average pool is used to fuse features and change the dimension for the final convolution. The length of video sequence is 300 and we set $N=4$ to increase the receptive filed. For convolutions, we set $C=512$ output channels to maintain a balance between accuracy and complicity and the dropout rate $p=0.25$.

\renewcommand{\multirowsetup}{\centering}
\begin{table}
\small
\begin{center}
\begin{tabular}{c|c|c} \hline

\rule{0pt}{10pt} \multirow{1}{1.5cm}{layer name}  & \multirow{1}{1.5cm}{output size}  & \multirow{1}{1.5cm}{layer} \\ \hline

\rule{0pt}{10pt} \multirow{1}{1.5cm}{conv\_1} & \multirow{1}{1.5cm}{$(512,298)$}   & \multirow{1}{1.5cm}{$3d1,512$} \\ \hline

\rule{0pt}{10pt} \multirow{2}{1.5cm}{res\_1} & \multirow{2}{1.5cm}{$(512, 292)$} & \multirow{2}{2cm}{$3d3,512$ \\ $1d1,512$} \\ 
\rule{0pt}{8pt} & & \\ \hline

\rule{0pt}{10pt} \multirow{2}{1.5cm}{res\_2} & \multirow{2}{1.5cm}{$(512, 274)$} & \multirow{2}{2cm}{$3d9,512$ \\ $1d1,512$} \\ 
\rule{0pt}{8pt} & & \\ \hline

\rule{0pt}{10pt} \multirow{2}{1.5cm}{res\_3} & \multirow{2}{1.5cm}{$(512, 220)$} & \multirow{2}{2cm}{$3d27,512$ \\ $1d1,512$} \\ 
\rule{0pt}{8pt} & & \\ \hline

\rule{0pt}{10pt} \multirow{2}{1.5cm}{res\_4} & \multirow{2}{1.5cm}{$(512, 58)$} & \multirow{2}{2cm}{$3d81,512$ \\ $1d1,512$} \\ 
\rule{0pt}{8pt} & & \\ \hline

\rule{0pt}{10pt} \multirow{1}{1.5cm}{pooling}  & \multirow{1}{1.5cm}{$(512)$}  & \multirow{1}{1.7cm}{$average \ pool$} \\ \hline

\rule{0pt}{10pt} \multirow{1}{1.5cm}{conv\_2}  & \multirow{1}{1.5cm}{$(2)$}  & \multirow{1}{1.5cm}{$1d1,2$} \\ \hline
\end{tabular}
\end{center}
\caption{This table shows the architecture and output size of each block for our fall detection network. "$3d3,512$" means 1D convolution with kernel size 3, dilation factor 3 and 512 channels.}
\label{size}
\end{table}

\begin{figure*}[ht]
\begin{center}
\includegraphics[scale=0.67]{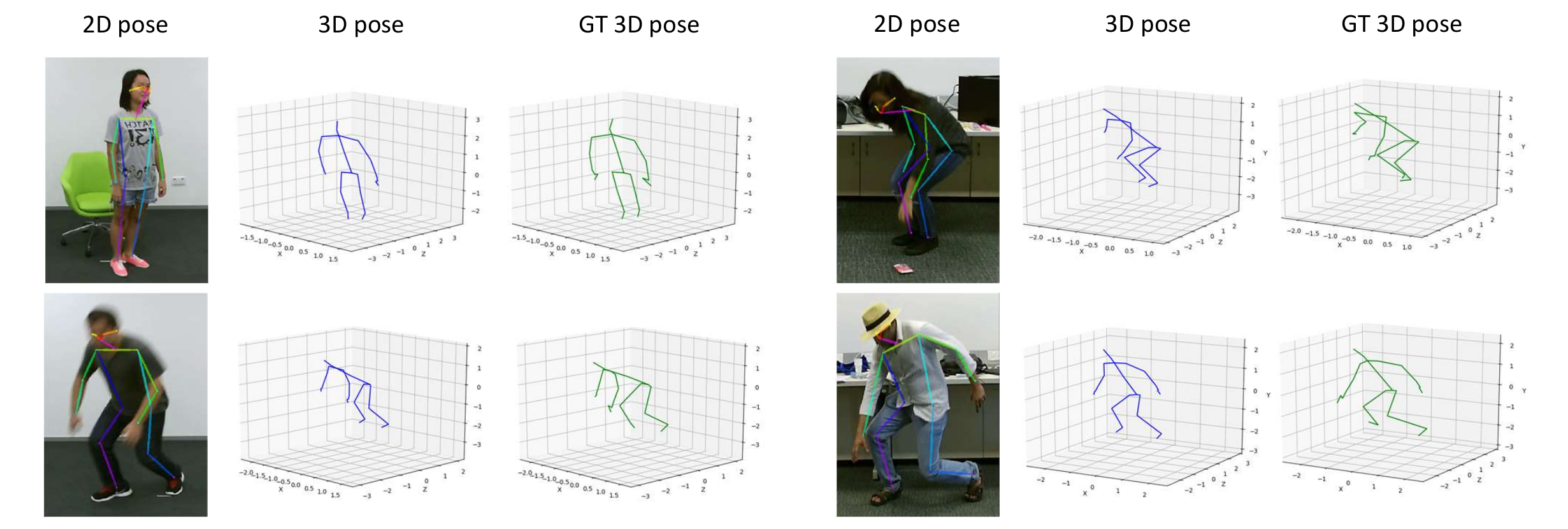}
\end{center}
\caption{Qualitative results of some example images. Initial image, 2D poses, 3D poses and GT 3D poses are presented. Blue poses are estimated 3D poses and green ones are GT 3D poses.}
\label{Results}
\end{figure*}

\section{Experiments}

\subsection{Dataset}
The proposed fall detection model was trained and evaluated on NTU RGB+D Action Recognition Dataset \cite{ntu} made available by the ROSE Lab at the Nanyang Technological University, Singapore. 
This dataset contains 60 action classes and 56,880 video samples including falling.
The videos are captured by three synchronous Microsoft Kinect v2 cameras installed at the same height with three different horizontal angles: $-45^{\circ}, 0^{\circ}, +45^{\circ}$. 
The dataset contains RGB videos, depth map sequences, 3D skeletal data, and infrared (IR) videos for each sample.
The resolution of RGB frames are $1920 \times 1080$ and the speed are 30 FPS. 
3D skeletal data are composed of 3D coordinates of 25 joints. 
To best of our knowledge, this is the largest action recognition dataset available that contains 3D skeletal data and falling samples. 
Some of the video samples have missing frames, poses or involve more than one person, we removed these samples.
Consequently, the total amount of samples we used was 44372, in which 890 samples were falling samples. 
Following previous action recognition work \cite{Action}, we trained on data coming from camera $0^{\circ}$ and $+45^{\circ}$, tested on data from camera $-45^{\circ}$.

\subsection{Training details}
We trained our fall detection model step by step. 
Firstly, for human pose estimation, we adopt off-the-shelf LPN to predict 2D poses from each video frame. 
LPN was pretrained on COCO dataset and fine-tuned on NTU RGB+D datase. 
When fine-tuning LPN on NTU RGB+D dataset, joint heatmaps were generated according to annotations as the output target which could avoid directly learning the mapping from images to coordinates. 
Then by calculating the center of mass of heatmaps, 2D joint coordinates could be obtained.

Before lifting 2D poses to 3D poses, 2D poses were normalized by centering to its root joint and then scaled by dividing its Frobenius norm. 
Adam optimizer was used to train the lifting network combined with MSE loss for 60 epochs with an initial learning rete of 0.0001 and exponential decay at $20th$ and $40th$ epoch.

Similar to the training of lifting network, 3D poses were normalized before being sent to the fall detection network.
Video samples from dataset have different frames, so all samples were expanded to 300 frames by padding null frames with previous ones. 
Adam optimizer and Cross Entropy Loss were used to train fall detection network. We set initial learning rate to 0.0001  with exponential decay. We train the network on one Nvidia GTX 1660 GPU for 20 epochs.

\subsection{Results}

\textbf{Human Pose Estimation} 
Most previous skeleton based action recognition works directly use 3D annotations on NTU RGB+D dataset. Here we use predicted 3D poses for fall detection to test feasibility of our approach. Human pose estimation accuracy is measured by Joint Detection Rate (JDR). 
JDR represents the percentage of successfully detected joints.
A joint is regarded as successfully detected if the distance between the estimation and groundtruth is smaller than a threshold. Here we set the threshold to be half of the distance between neck and head.

\begin{figure}[ht]
\begin{center}
\includegraphics[scale=0.5]{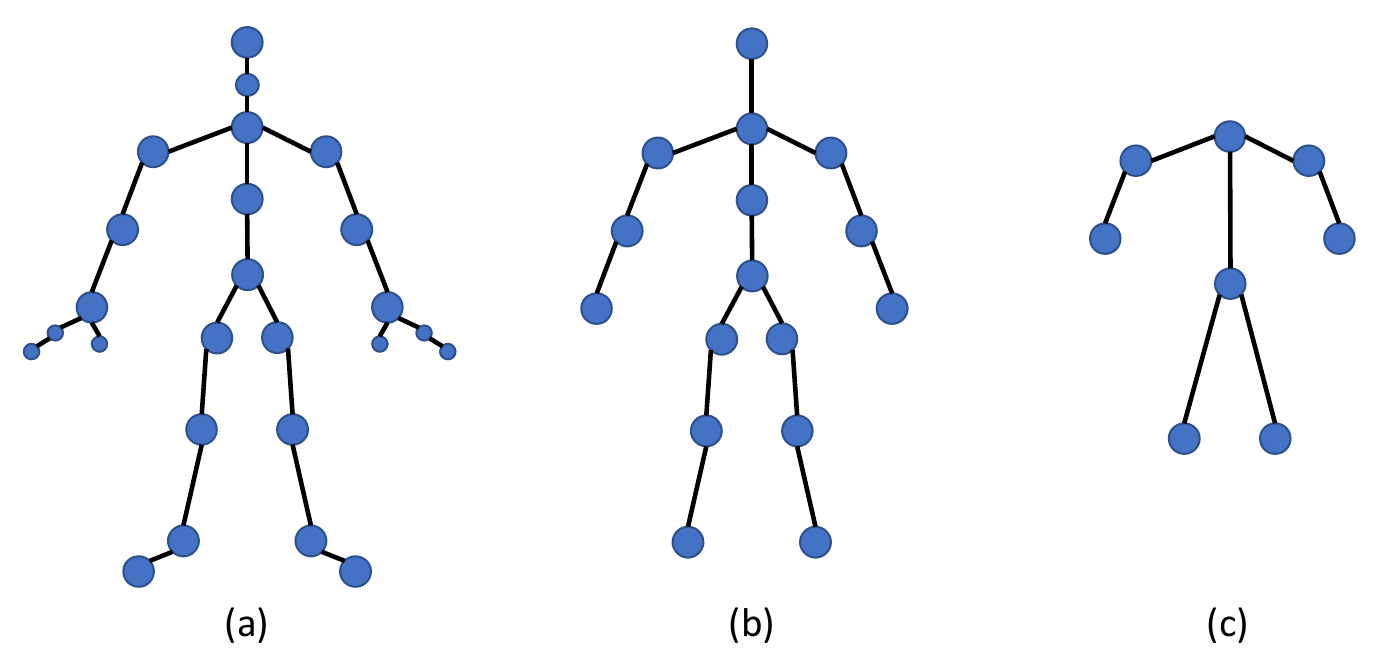}
\end{center}
\caption{Skeleton information of different inputs. (a) All 25 joints. (b) Selected 16 joints. (c) Selected 8 joints.}
\label{joints}
\end{figure}

Table \ref{JDR1} shows pose estimation results of our method on NTU RGB+D dataset. JDR of some joints are presented. For some joints including head, elbow, shoulder, JDRs achieve larger than $90\%$ and the prediction of these joints is accurate. However, JDR of ankle and thumb are not high as a result of frequent occlusion and inaccurate 2D pose. Table \ref{JDR2} shows mean JDR of three aggregations of joints. Details of three aggregations of joints are showed in Figure \ref{joints} The evaluation of our proposed fall detection method using different number of joints as inputs will later be reported.

\begin{table}[ht]
\small
\begin{center}
\begin{tabular}{ccccccc}
\hline
\rule{0pt}{8pt}B spi & Head  & L elb  & L wri & R elb  & R wri & R ank \\ \hline
\rule{0pt}{8pt}99.69 & 98.02 & 98.45  & 97.91 & 94.22  & 90.82 & 71.06 \\ \hline
\end{tabular}
\end{center}
\caption{This table shows pose estimation accuracy on NTU RGB+D dataset. JDR (\%) of seven joints are showed due to limited space. "B spi" means base spine and "L wri" means left wrist.}
\label{JDR1}
\end{table}

\begin{table}[ht]
\small
\begin{center}
\begin{tabular}{cc}
\hline
Joints & mJDR \\ \hline
\rule{0pt}{9pt} 25 joints                  & 86.02                    \\
\rule{0pt}{9pt} 16 joints                  & 94.07                    \\
\rule{0pt}{9pt} 8 joints                   & 94.52                    \\ \hline
\end{tabular}
\end{center}
\caption{This table shows mean JDR (\%) of three aggregations of joints.}
\label{JDR2}
\end{table}

\begin{table}[ht]
\small
\begin{center}
\begin{tabular}{ccccc}
\hline
\rule{0pt}{9pt} Methods  & Input  & Feature & Network & Accuracy \\ \hline
\rule{0pt}{9pt}  Xu et al. \cite{F4} &  RGB  &  Pose &  2D conv & 91.70\% \\ 
\rule{0pt}{9pt}  Anahita et al. \cite{F7}  &  Depth  &  Pose &  LSTM & 96.12\%\\ 
\rule{0pt}{9pt}  Han et al. \cite{F8} &  Depth  &  Pose &  1D conv & 99.20\%\\
\rule{0pt}{9pt}  \textbf{Ours}  &  RGB  &  Pose &  1D conv & \textbf{99.83}\%\\ \hline
\end{tabular}
\end{center}
\caption{Fall detection accuracy of different methods on NTU RGB+D dataset. }
\label{compare}
\end{table}

\begin{table}[ht]
\small
\begin{center}
\begin{tabular}{cccc}
\hline
\rule{0pt}{9pt} Method          & Accuracy         & Precision        & Recall \\ \hline
\rule{0pt}{9pt} 8 joints-CEL   & 99.72\%          & 97.15\%          & 89.74\%       \\
\rule{0pt}{9pt} 8 joints-WCEL  & 99.29\%          & 98.70\%          & 74.32\%       \\
\rule{0pt}{9pt} 16 joints-CEL  & \textbf{99.83\%} & 97.47\%          & \textbf{94.25\%} \\
\rule{0pt}{9pt} 16 joints-WCEL & 99.50\%          & \textbf{98.73\%} & 80.67\%       \\
\rule{0pt}{9pt} 25 joints-CEL  & 99.77\%          & 97.79\%          & 91.35\%       \\
\rule{0pt}{9pt} 25 joints-WCEL & 99.66\%          & 97.47\%          & 87.11\%       \\ \hline
\end{tabular}
\end{center}
\caption{ Fall detection results of different methods on NTU RGB+D dataset. The naming convention of the methods follows the rule of “A-B” where “A” indicates how many joints are used in fall detection. "B" denotes the loss. "CEL" means Cross Entropy Loss and "WCEL" means weighted Cross Entropy Loss. }
\label{loss}
\end{table}

\textbf{fall detection} Table \ref{compare} shows the result of our proposed method and other fall detection methods. It can be seen that our proposed method achieve a quite high performance of $99.83\%$ accuracy and outperforms other methods. 

We also evaluate the influence of the number of input joints, as shown in Figure \ref{joints}, we select 8 and 16 joints from all the 25 joints as input and calculate the classification accuracy. Table \ref{loss} shows the results of different joint input. It can be seen that when using 16 joints as input, the model achieves the highest accuracy of $99.83\%$. Using all the 25 joints or 8 joints achieve a lower but still high accuracy. We consider that some joints like eyes, hands could disturb the model to learn action features, which could lead to the degradation of accuracy. Besides, few joints may not be able to model the variance of different actions. 

Considering that the number of falling samples only takes a part of 1.67 percentage of the whole dataset, it is necessary to test whether our model can truly classify fall from other actions rather than just classifying all the samples to not fall class. So Precision is also calculated to test whether our model can really recognize fall behaviour. Moreover, we use Weighted Cross Entropy Loss to train this fall detection network follow previous method and evaluate it on this dataset. For falling class, we set $\alpha=59/60$ and $\beta=1/60$ for other samples. Table \ref{loss} shows the evaluation results, we can see that our network truly learns how to classify a fall behavior and the precision achieves $1.26\%$ improvement using Weighted Cross Entropy Loss while accuracy decreased by $0.83\%$. The variation of Recall is very large as the number of falling samples is much smaller than it of other actions.   

Actual inference speed is what we care about which defines whether our fall detection method can achieve real time. We test number of parameters, FLOPs and inference speed of every part of our fall detection approach. The speed test is based on two platforms. The one is a non-GPU platform with Intel Core i5-9400F CPU (2.9GHZ$\times$6) and the other is one Nvidia GTX 1660 GPU. Table \ref{speed} shows the measurements and we can find that our method has a speed of 18 FPS on a non-GPU platform and 63 FPS on one Nvidia GTX 1660 GPU. Moreover, the inference speed of lifting network and fall detection network is very fast that only takes a few milliseconds. The 2D pose estimator, LPN, is the one mainly limits the inference speed. It is worth mentioning that our fall detection method does not rely on any specific 2D human pose estimator, LPN can be changed to other pose estimator for more efficiency and robustness.

\begin{table}
\small
\begin{center}
\begin{tabular}{c|c|c|c|c}
\hline
\rule{0pt}{9pt} Part & Params & FLOPs & non-GPU & GPU \\ \hline
\rule{0pt}{9pt} LPN     & 2.7M   & 1.0G  & 20\ FPS & 74\ FPS \\ \hline
\rule{0pt}{9pt} LN        & 2.2M & 0.28G & 560\ FPS & 1450\ FPS \\ \hline
\rule{0pt}{9pt} FDN & 4.2M & 0.9G  & 260\ FPS & 590\ FPS \\ \hline
\rule{0pt}{9pt} Whole   & 9.1M  & 2.18G  & 18\ FPS & 63\ FPS  \\ \hline
\end{tabular}
\end{center}
\caption{ Measurement of Params, FLOPs and Speed of each part of our proposed fall detection method on different platforms. "LN" means lifting network and "FDN" means fall detection network.}
\label{speed}
\end{table}

\section{Conclusion}
In this paper, we propose an approach to recognize fall events from video sequences. More specifically, our approach includes a 3D pose estimator based on lifting 2D poses to 3D poses and a fall detection network using dilated convolution. Our approach achieves a high accuracy of 99.83 on NTU RGB+D dataset and realtime performance of 18 FPS on a non-GPU platform and 63 FPS on a GPU platform.

{\small
\bibliographystyle{ieee_fullname}
\bibliography{egbib}
}

\end{document}